\documentclass[conference]{IEEEtran}
\pdfoutput=1
\usepackage{cite}
\usepackage{amsmath,amssymb,amsfonts}
\usepackage{algorithm}
\usepackage{algorithmic}
\usepackage{graphicx}
\usepackage{textcomp}
\usepackage{xcolor}
\def\BibTeX{{\rm B\kern-.05em{\sc i\kern-.025em b}\kern-.08em
    T\kern-.1667em\lower.7ex\hbox{E}\kern-.125emX}}

\usepackage{CJKutf8}
\usepackage{multirow}       
\usepackage{subcaption}     
\usepackage{makecell}       
\usepackage{booktabs}       
\usepackage{url}
\usepackage{xpinyin}
\newcommand{\cjksong}[1]{{{\begin{CJK*}{UTF8}{gbsn}#1\end{CJK*}}}}

\begin{document}

\title{
Make BERT-based Chinese Spelling Check Model Enhanced by Layerwise Attention and Gaussian Mixture Model
}

\author{\IEEEauthorblockN{1\textsuperscript{st} Yongchang Cao}
\IEEEauthorblockA{\textit{National Key Laboratory for Novel Software Technology} \\
\textit{Nanjing University}\\
Nanjing, China \\
caoyc@smail.nju.edu.cn}
\and
\IEEEauthorblockN{2\textsuperscript{nd} Liang He}
\IEEEauthorblockA{\textit{National Key Laboratory for Novel Software Technology} \\
\textit{Nanjing University}\\
Nanjing, China \\
heliang@smail.nju.edu.cn}
\and
\IEEEauthorblockN{3\textsuperscript{rd} Zhen Wu}
\IEEEauthorblockA{\textit{National Key Laboratory for Novel Software Technology} \\
\textit{Nanjing University} \\
Nanjing, China \\
wuz@nju.edu.cn}
\and
\IEEEauthorblockN{4\textsuperscript{th} Xinyu Dai*}
\IEEEauthorblockA{\textit{National Key Laboratory for Novel Software Technology} \\
\textit{Nanjing University} \\
Nanjing, China \\
daixinyu@nju.edu.cn}
}

\maketitle

\begin{abstract}
BERT-based models have shown a remarkable ability in the Chinese Spelling Check (CSC) task recently. However, traditional BERT-based methods still suffer from two limitations. First, although previous works have identified that explicit prior knowledge like Part-Of-Speech (POS) tagging can benefit in the CSC task, they neglected the fact that spelling errors inherent in CSC data can lead to incorrect tags and therefore mislead models. Additionally, they ignored the correlation between the implicit hierarchical information encoded by BERT's intermediate layers and different linguistic phenomena. This results in sub-optimal accuracy. To alleviate the above two issues, we design a heterogeneous knowledge-infused framework to strengthen BERT-based CSC models. To incorporate explicit POS knowledge, we utilize an auxiliary task strategy driven by Gaussian mixture model. Meanwhile, to incorporate implicit hierarchical linguistic knowledge within the encoder, we propose a novel form of n-gram-based layerwise self-attention to generate a multilayer representation. Experimental results show that our proposed framework yields a stable performance boost over four strong baseline models and outperforms the previous state-of-the-art methods on two datasets.
\end{abstract}

\section{Introduction}
Chinese Spelling Check (CSC) is the crucial task of detecting and correcting spelling errors contained in a given input. Spelling errors are common in daily life. Designing a high-quality spelling checker can serve many NLP tasks, such as improving the quality of ASR and OCR~\cite{afli2016using}, used for essay scoring~\cite{burstein-chodorow-1999-automated}, or more downstream NLP tasks, such as search engines~\cite{gao2010a}.

Earlier work on CSC followed a rule-driven pipeline of error detection, candidate generation, and candidate selection~\cite{liu2013hybrid, yu2014chinese}. With the development of neural networks, some works attempted to solve the CSC problem using sequence-to-sequence models~\cite{wang-etal-2019-confusionset}. Most recently, BERT-based models~\cite{devlin2019bert} have been introduced into the CSC task and achieved state-of-the-art (SOTA) performance. Most BERT-based CSC models can be grouped into two categories: introducing separate error detection modules~\cite{zhang-etal-2020-spelling, zhang2021correcting} and introducing similarity constraints through phonetic and glyph information~\cite{cheng-etal-2020-spellgcn, xu-etal-2021-read}.

Despite BERT's powerful natural language understanding ability, the existing BERT-Based CSC models are still limited by inadequate utilization of versatile linguistic knowledge, thus hindering the potential of the models for spelling error correction. Concretely, on the one hand, explicit Part-Of-Speech (POS) knowledge as an input feature has been proven to be beneficial in detecting spelling errors~\cite{yang2017alibaba, fu2018chinese, Xie2019AutomaticCS, cao-etal-2020-integrating}. However, spelling errors in the CSC corpus will result in severe incorrect labeling by the publicly available POS tagging tools trained on the clean corpus, and the way of simply inputting POS features cannot avoid the misleading of noisy labels for the CSC model. On the other hand, previous works have shown that BERT comprises various types of linguistic features~\cite{jawahar:2019}, and different linguistic features are adept at handling different types of errors~\cite{li:2021}. Existing BERT-based CSC methods solely use the top-layer latent representation to make corrections, ignoring the assistance of various linguistic features for Chinese spelling errors, hence weakening the performance of CSC models. Our preliminary investigations demonstrate that low-level information supplementation can indeed yield higher detection and correction performance for non-word type errors\footnote{Spelling errors can be divided into non-word type that causes lexical anomalies and real-word type that causes semantic anomalies~\cite{Xie2019AutomaticCS, wang2021chinese}.}. Nevertheless, effective interlayer information fusion strategies specific to the type of spelling error remain to be explored.

To address the above two issues, we design a practical heterogeneous knowledge-infused framework, i.e., explicit POS knowledge and implicit hierarchical linguistic features in BERT, to strengthen the BERT-based CSC models. Specifically, to solve the first issue aforementioned, we utilize an auxiliary-task learning strategy driven by Gaussian Mixture Model (GMM). We perform token-level and task-level loss annealing by the GMM and heuristic strategy for the noisy POS feature, while using the auxiliary task to fuse explicit knowledge, the utilized loss annealing strategy can effectively reduce the underlying interference of massive noisy POS tags to the CSC model. For the second limitation, we exploit a layerwise self-attention mechanism based on n-gram tokens to establish an information pipe between the intermediary encoder layers and the classifier. The exploited n-gram token-based attention query term can provide a well-focused information supplement specific to the type of error.

We employ our proposed framework on four strong baseline models. Results from extensive experiments indicate that our proposed framework can yield a stable performance improvement on the BERT-based CSC models and outperform the previous state-of-the-art method across two datasets. Further in-depth analysis also validates the effectiveness of each proposed module. The contributions are summarized as follows: 

\begin{itemize}
    \item A new framework named Auxiliary Task learning based on Loss Annealing with Layerwise self-Attention (ATLAs) is designed to improve BERT-based CSC models, which can be universally and effectively applied to the diverse array of BERT-based CSC models.
    \item A loss annealing strategy is utilized for auxiliary task training, which reduces the sensitivity of baseline models to the massive incorrect POS labeling and alleviates the performance degradation in knowledge-dependent Chinese spelling error correction.
    \item Several multilayer representation techniques have been explored and compared, and the exploited n-gram-based layerwise self-attention is verified to be effective in the CSC model.
    \item Extensive experiments applied to four strong BERT-based CSC models show that our knowledge-infused framework can improve all the baseline models across two different datasets and achieve state-of-the-art performance for the CSC task.
\end{itemize}

\section{Related Works}
\label{sec:related}

\subsection{BERT-Based CSC Models}
As a critical NLP application, The CSC task has attracted much attention from the NLP community. Due to the context-sensitive nature of the CSC task~\cite{wang-etal-2019-confusionset}, recent models all use BERT as the base corrector. The BERT-based corrector mainly has two optimization strategies. One is to add an independent detection module. Soft-Masked BERT~\cite{zhang-etal-2020-spelling} leverages a Bi-GRU network to locate spelling errors and uses the error probability to soft embed input characters. The other is to introduce the phonetic and graphic information of characters. E.g., SpellGCN~\cite{cheng-etal-2020-spellgcn} and ReaLiSe~\cite{xu-etal-2021-read} use fixed similar character sets or pre-trained models to incorporate visual and phonological similarity knowledge into the CSC model to modify the prediction logits of the Masked Language Model. However, these models ignore the dependence of the CSC task on explicit prior knowledge of POS and implicit hierarchical linguistic features in BERT.

\subsection{Explicit Prior Knowledge Injection}
Prior knowledge has been introduced in many knowledge-intensive NLP tasks to enhance the performance of the pre-trained model~\cite{he-etal:2020, Xia:2021}. DPL-Corr~\cite{Xie2019AutomaticCS} combines the POS tag with contextual character representation and uses the mixed information for the detection module in the CSC pipeline framework. Reference~\cite{li-etal-2021-exploration} continuously identifies weaknesses during the CSC model training process and generates adversarial samples to incorporate explicit knowledge that may be lacking in the BERT model. However, these methods either inevitably introduce noise in incorrect POS tags or increase the cost of training. In contrast, our strategy does not require additional training costs and reduces the impact of severe knowledge noise.

\subsection{Implicit Hierarchical Features in BERT} 
Many analytical or practical works have demonstrated that BERT composes hierarchical linguistic features at different layers. Reference~\cite{jawahar:2019} verified that BERT captures phrase-level information in the lower layers, and this information gradually dilutions in the higher layers. BERT4GCN~\cite{xiao-etal-2021-bert4gcn} utilizes outputs from intermediate layers of BERT and positional information to augment Graph Convolutional Network and verifies the benefits of the strategy in the aspect-based sentiment classification task. Most relevant to the CSC task,~\cite{li:2021} revealed that out-of-domain points in different layers correspond to different linguistic phenomena, e.g. lower layers correspond to low-frequency tokens. In comparison, grammatically anomalous inputs are out-of-domain in higher layers. However, to the best of our knowledge, no attempt has been made to investigate how to integrate well-focused hierarchical information to assist in spelling error correction.

\section{Methodology}
\label{sec:model}

\subsection{Problem Formulation}

The Chinese Spelling Check task can be formalized as follows: Given a Chinese sequence of $n$ characters $X = \{x_1, x_2, \dots, x_n\}$, the goal of the model is to convert it into a sentence $\hat{Y} = \{\hat{y_1}, \hat{y_2}, \dots, \hat{y_n}\}$, where the misspelled characters in $X$ will be replaced with the correct characters in $\hat{Y}$. In the public Chinese Spelling Check shared task, $X$ and $\hat{Y}$ are set to have the same length. The CSC model function can be regarded as a mapping function $f:X \rightarrow \hat{Y}$. Compared to traditional translation tasks, most of the characters in $\hat{Y}$ in the CSC task are copied directly from $X$.

\subsection{Generic BERT-Based CSC Model}
\begin{figure}[t]
  \centering
  \includegraphics[width=0.65\columnwidth]{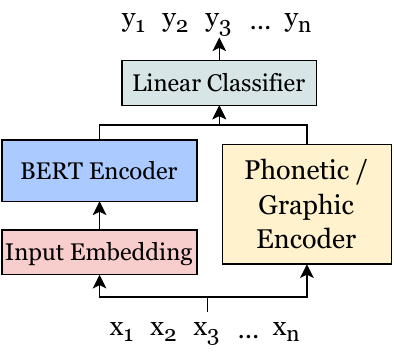}
  \caption{The generic framework of BERT-based CSC models, which usually includes a BERT-based character encoder and a phonetic and graphic encoder.}
  \label{fig_baselines}
\end{figure}

Most recent BERT-based CSC models employ BERT as the character feature extractor and utilize a unique phonetic and graphic encoder to fuse additional similarity constraints for spelling correction. A general model framework is shown in Fig.~\ref{fig_baselines}. Input $X$ obtains the character embedding $\boldsymbol{E} = \{\mathbf{e}_1, \mathbf{e}_2, \dots, \mathbf{e}_n\}$ through the BERT embedding layer, and then obtains the contextual representation $\{\mathbf{h}^L_1, \mathbf{h}^L_2, \dots, \mathbf{h}^L_n\}$ through 12 transformer blocks, where $L$ represents the number of encoder layers in BERT. In addition, different strategies are adopted to obtain the phonetic and glyph information. Ultimately, the contextual representation with additional information will be sent to a FullyConnected classifier to predict the correct sequence $\hat{Y}$.

\subsection{N-gram Based Layerwise Self-Attention}
\label{subsec:layerwise}
\begin{figure*}[t]
  \centering
  \includegraphics[width=0.7\textwidth]{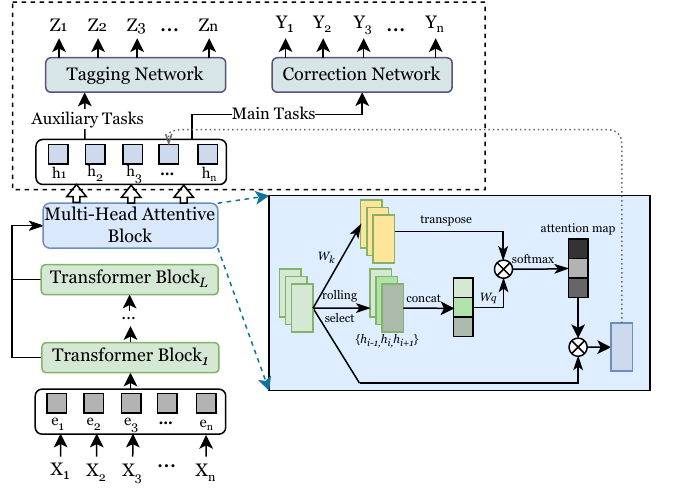}
  \caption{The framework ATLAs we proposed. It contains a layerwise self-attention module and an auxiliary task joint learning module.}
  \label{fig_MTLLA}
\end{figure*}

Many probing tasks have proven that different BERT layers have individual abilities to uncover linguistic anomalies~\cite{li:2021}. Our preliminary experiments also validate that varied levels of intermediate information within BERT are good at handling different types of spelling errors. To take full advantage of the implicit hierarchical knowledge inside the BERT encoder, we exploit a strategy of initiating an attention query on the encoder's intermediary layers by using n-gram tokens to fuse richer information. The primary processing step is shown on the right side of Fig.~\ref{fig_MTLLA}.

Specifically, for each input character $x_i$, assume that its BERT middle-level encoding representations are symbolized as $\boldsymbol{H}_i = \{\mathbf{h}_i^0, \mathbf{h}_i^1, \dots, \mathbf{h}_i^L\}$ (where $\mathbf{h}_i^0$ is the embedding $\mathbf{e}_i$), layerwise self-attention calculates the final representation $\widehat{\mathbf{h}}_i$ of the character $x_i$ as follows:

\begin{equation}
  \begin{aligned}
    \boldsymbol{q}_i &= [\mathbf{h}_{i-1}^L; \mathbf{h}_{i}^L; \mathbf{h}_{i+1}^L] * \boldsymbol{W}_Q \\
      \boldsymbol{K}_i &= \boldsymbol{H}_i * \boldsymbol{W}_K \\
      S_i &= \textrm{softmax} (\frac{\boldsymbol{q}_i \cdot \boldsymbol{K}_i^\top}{\sqrt{d_{K}}}) \\
      \widehat{\mathbf{h}}_i &=  \sum_{\ell = 0}^{L} s_i^\ell \cdot \mathbf{h}_i^\ell  
  \end{aligned}
  \label{equa_LA}
\end{equation}

\noindent Where $i \in [1, n] $ represents the character positional index of the input $X$, $\boldsymbol{W}_Q, \boldsymbol{W}_K$ are trainable matrices that generate the query vector $\boldsymbol{q}_i$ and the key array $\boldsymbol{K}_i$ of the character $x_i$, and $\boldsymbol{K}_i$ corresponds to $\boldsymbol{H}_i$, which has $L+1$ vectors, representing the number of encoder layers. $[\cdot]$ is the vector concatenation operation to generate the n-gram token. Then, the mixed representation $\boldsymbol{q}_i$ is used to query the dependence of the character $x_i$ on low-level information according to the n-gram form token at positions $i$. Finally, the resulting attention score $S_i$ is used to calculate a weighted sum of the multilayer representations. Essentially, $S_i$ represents the supplemented information achieved from the low-level to top-level representation. Note that we omit the activation matrix for the value vectors to retain the original semantics learned by the BERT. Owing to the query based on n-gram representation, the strategy can fine-tune the preferenced intermediary representations based on the different spelling error types.

We also use the multi-head mechanism. Specifically, for the $m$-head self-attention, the attention operation first divides the original latent embedding into $m$ subspaces, reducing the computational cost and making the model pay attention to different aspects of information. For the $j$-th attention header, it corresponds to the subspace $\widetilde{\boldsymbol{Q}}_j, \widetilde{\boldsymbol{K}}_j, \widetilde{\boldsymbol{V}}_j$, and uses~\eqref{equa_LA} to calculate self-attention. Lastly, all heads are concatenated to obtain the ultimate characters latent representation $\widehat{\boldsymbol{H}} = \{\widehat{\mathbf{h}}_1, \widehat{\mathbf{h}}_2, \dots, \widehat{\mathbf{h}}_n\}$. The operation can be formulated as follows:

\begin{gather}
  \label{equa_MHLA}
  \begin{split}
    head_j = Attention&(\widetilde{\boldsymbol{Q}}_j, \widetilde{\boldsymbol{K}}_j, \widetilde{\boldsymbol{V}}_j) \\
    MultiHead(\boldsymbol{Q}, \boldsymbol{K}, \boldsymbol{V}) =& \; [head_1, head_2, \dots, head_m] \\
  \end{split}
\end{gather}

\subsection{Auxiliary Loss Annealing Joint Learning}
\label{subsec:Auxiliary}
Explicit POS information provides good guidance for identifying many commonly-confused Chinese characters~\cite{Xie2019AutomaticCS}. However, the misspellings contained in the corpus guarantee that severe noise is contained in the knowledge annotations based on the publicly available tagging tools. How to mitigate the impact of noise labeling in knowledge injection is the key to the CSC task. For this reason, we designed an auxiliary-task joint-learning strategy based on loss annealing. Specifically, after obtaining the multilayer representation $\widehat{\boldsymbol{H}}$, the model utilizes separate plain FullyConnected(FC) layers as classifiers to make predictions for the main CSC task as well as the auxiliary POS tagging task for knowledge injection. After obtaining the spelling prediction result $Y = \{y_1, y_2, \dots, y_n\}$ and the POS tagging result $Z = \{z_1, z_2, \dots, z_n\}$, the model calculates the corresponding losses of main task $\mathcal{L}^m$ and the auxiliary task $\mathcal{L}^a$ through the relevant labels as shown in \eqref{equa_Loss_pre:eq2}.

\begin{equation}
\label{equa_Loss_pre:eq2}
\begin{split}
    \mathcal{L}^a \; &= \; -\sum\nolimits_{i=1}^n \mathcal{L}^a_i \; = \; -\sum\nolimits_{i=1}^n \log{p_a(z_i = \hat{z_i} | X)} \\
    \mathcal{L}^m \; &= \; -\sum\nolimits_{i=1}^n \mathcal{L}^m_i \; = \; -\sum\nolimits_{i=1}^n \log{p_m(y_i = \hat{y_i} | X)} \\
\end{split}
\end{equation}

\noindent Where $\hat{Z}$ represents the prior label for auxiliary task. Since the losses of noisy labels and clean labels tend to be subject to different Gaussian distributions~\cite{zhang2021understanding, qiao2022selfmix}, after acquiring the $\mathcal{L}^m$ of the auxiliary task, we apply the popular used GMM~\cite{arazo2019unsupervised} to distinguish noisy labels by feeding the auxiliary loss items $\mathcal{L}^m$. We feed the auxiliary losses to a 2-component Gaussian distribution and use Expectation-Maximization (EM) algorithm\footnote{\url{https://en.wikipedia.org/wiki/Expectation-Maximization}} to fit the GMM to the observations. Let $\alpha_{i}$ represent the probability of $i^{th}$ pos tag belonging to the Gaussian component with the smaller mean, which can also be considered as the clean probability due to the small-loss theory~\cite{arpit2017closer}. The final loss function in the forward propagation is calculated as~\eqref{equa_Loss:eq2}:

\begin{equation}
\label{equa_Loss:eq2}
\begin{split}
    \mathcal{L} &= \eta_t * \mathcal{L}_a + \alpha_i \times (1 - \eta_t) * \mathcal{L}_m  \\
    \eta_t &= \frac{1}{1+e^{\beta(-t+\frac{T}{2})}}
\end{split}
\end{equation}

\noindent Where $\eta$ represents the annealing factor, which is in the form of inverse sigmoid, $t$ indicates the current forward step, and $T$ indicates the total number of updates in the training phase, $\beta$ is the smoothing factor.

The auxiliary tagging task essentially injects explicit POS information into the BERT representation by fine-tuning the pretrained language model, and the annealing strategy provides different sensitivity to prior knowledge at different characters and different stages in the training process. 

Finally, we summarize the above training process as Algorithm \ref{alg:A}.
\begin{algorithm}  
\caption{Algorithm for Training Phase of ATLAs Framework}  
\label{alg:A}  
\begin{algorithmic}[1]
\REQUIRE{Traing Set $\mathcal{D}$}
\ENSURE{The Chinese spelling checker $f$}
\STATE {Initialize GMM using auxiliary task losses generated by initialization parameters}
\FOR {$epoch = 1$ \TO $EPOCHS$}
    \STATE {$\mathcal{L}^m \gets \{\}$}
    \FOR {$i = 1$ \TO $|\mathcal{D}|$}
        \STATE {Randomly sample a batch of data $B_x$}
        \STATE {$H_x \gets$  Encoding input $B_x$ using the BERT model}
        \STATE {$\hat{H_x} \gets$ Using Equation \eqref{equa_LA} to perform Layerwise self-Attention on $H_x$}
        \STATE {Calculate $\mathcal{L}^a_x$ and $\mathcal{L}^m_x$ according to Equation \eqref{equa_Loss_pre:eq2}}
        \STATE {Estimate $\alpha_x$ based on GMM}
        \STATE {Update $f$ according to Equation \eqref{equa_Loss:eq2}}
        \STATE {$\mathcal{L}^m \gets \mathcal{L}^m + \mathcal{L}^m_x$}
    \ENDFOR
    \REPEAT \STATE {Update GMM model using EM algorithm}
    \UNTIL{The GMM model converges to the observation $\mathcal{L}^m$}
\ENDFOR
\end{algorithmic}  
\end{algorithm}

\section{Experiment}
\label{sec:experiment}

We use the same data preprocessing method as previous works \cite{wang-etal-2019-confusionset, cheng-etal-2020-spellgcn, xu-etal-2021-read}. Specifically, to verify the practicality of the proposed ATLAs strategy, we use two manual CSC datasets released by SIGHAN \cite{yu-etal-2014-overview, tseng-etal-2015-introduction}. As with the SpellGCN and ReaLiSe model, we also introduced the additional 271K generative corpus \cite{wang-etal-2018-hybrid} to ensure comparability. We discarded the 2013 dataset due to the poor annotation quality, for which a good-performing model may obtain bad scores. Following previous works, the characters are converted to simplified Chinese using OpenCC\footnote{\url{ https://github.com/BYVoid/}}. The statistics of the data are displayed in Table~\ref{table_dataset}.

\begin{table}[ht]
  \centering
    \caption{Statistical information of the used dataset resources. The value in the Errors Line represents the number of sentences containing errors and the total number of sentences.}
  \label{table_dataset}
  {
  \begin{tabular}{l|c|c}
      \hline
      \textbf{Dataset} & \textbf{Errors Line} & \textbf{Avg. Length} \\
      \hline
      Wang271K & 271,009/271,329 & 44.4 \\
      2013 Train & 350/350 & 49.2 \\
      2014 Train & 2,432/3,437 & 49.6 \\
      2015 Train & 2,339/2,339 & 31.3 \\
      \hline
      Total & 277,130/277,455 & 44.4 \\
      \hline
      \hline
      2014 Test & 526/1062 & 50.1 \\
      2015 Test & 550/1100 & 30.5 \\
      \hline
  \end{tabular}
  }
\end{table}

\subsection{Baselines and Settings}

We apply our designed ATLAs to several baseline models to explore its universality and effectiveness. Since different CSC models employ different strategies, we make necessary modifications to apply the proposed optimization strategies. In each BERT-based CSC model, ATLAs is applied as follows:

\begin{itemize}
    \item \textbf{BERT-Finetune} is a model for fine-tuning BERT directly using the CSC task. We use the last layer of the BERT encoder to initiate the n-gram query to obtain a multilayer representation as described in section~\ref{subsec:layerwise}, and use the result to perform the CSC and auxiliary tasks.
    \item \textbf{Soft-Masked BERT}~\cite{zhang-etal-2020-spelling} proposes a soft-masking technique that sets the embedding of predicted misspelled characters to be similar to the \texttt{[MASK]} embedding. We only apply our strategies in its correction module as in the previous mode in the BERT-Finetune, without changing its principal detection module and soft-masking framework.
    \item \textbf{SpellGCN}~\cite{cheng-etal-2020-spellgcn} incorporates phonological and visual similarity knowledge into the BERT model through a specialized graph convolutional network on the fixed connected graph of similar characters. We use a plain application form of ATLAs, like in BERT-Finetune.
    \item \textbf{ReaLiSe}~\cite{xu-etal-2021-read} utilizes three additional Transformer Blocks to fuse multimodal information of the Chinese characters, including phonetic information encoded by Recurrent Neural Network and graphic information encoded by Convolutional Neural Network. We extend the layerwise self-attention range to ReaLiSe's additional transformer blocks and use the top-level representation to initiate the n-gram-based query. Other settings remain unchanged.
\end{itemize}

For all baseline models, we initialize BERT with the BERT-wwm model~\cite{cui-etal-2020-revisiting}. Following the original papers, we fine-tune ReaLiSe with the AdamW \cite{loshchilov2018fixing} optimizer for 10 epochs. For the remaining models, we fine-tune them with the AdamW optimizer for 6 epochs. The batch size is set to 64, and the learning rate is set to decay from 5e-5 to 1e-5 using the \textit{CosineScheduler}. The attention head $m$ is set to 8, and the smoothing factor $\beta$ is set to 8e-4. 
We use trigram to take into account adjacent characters on both sides. For the remaining parameters, we retain the settings of the original papers.

We recall that our proposed approach is approximately orthogonal to these based methods, which means that the original incorporation of phonetic or graphic information is preserved in our model wholly intact. 
It is worth noting that ATLAs neither increases the size of the training corpus nor the training epochs of the models, i.e., compare the related works that need to introduce additional training epochs \cite{li-etal-2021-exploration, xu-etal-2021-read}, there is no additional training cost.

For POS information, we use the THULAC tool\footnote{ \url{http://thulac.thunlp.org/}} to perform word segmentation and POS tagging on the training dataset. We also merge the location information into the feature itself, e.g., B-pos indicates the first character for a particular POS tag, while I-pos indicates the middle and end characters in a word.

\subsection{Main Results}

\begin{table*}[t]
  \caption{The performance of baseline models and ATLAs (\%), the results of the baseline models are derived from our reproduction results, which are close to the original paper. The best results are in bold.}
  \label{table_Results}
  \centering
  \setlength{\tabcolsep}{3mm}
  {
  \begin{tabular}{c|l|c|c|c|c|c|c|c|c}
      \hline
      \multirow{2}*{\textbf{Test Set}} & \makecell[c]{\multirow{2}*{\textbf{Method}}} &\multicolumn{4}{c}{\textbf{Detection-Level}} & \multicolumn{4}{|c}{\textbf{Correction-Level}} \\
      \cline{3-10}  
      & & \textbf{Acc.} & \textbf{Prec.} & \textbf{Rec.} & \textbf{F1.} & \textbf{Acc.} & \textbf{Prec.} & \textbf{Rec.} & \textbf{F1.} \\
      \hline
      \multirow{8}*{SIGHAN 2014} & BERT-Finetune & 76.8& 64.8& 67.9& 66.3& 76.0& 63.1& 66.1& 64.6 \\
      & BERT-ATLAs& 78.2& 67.0& 71.5& \textbf{69.2}& 77.4& 65.6& 70.0& \textbf{67.7} \\
      \cline{2-10}
      
      & Soft-Masked BERT&77.4	&65.7&	69.5&	{67.5}&	76.4&	63.6&	67.4&	{65.4}\\
      & SMB-ATLAs	& 78.9& 68.3& 70.0& \textbf{69.1}& 78.2& 66.8& 68.5& \textbf{67.6}\\
      \cline{2-10}

      & SpellGCN& 77.7&	66.2&	69.4&	{67.8}&	76.7&	64.4&	67.5&	{65.9}\\
      & SpellGCN-ATLAs& 79.1& 68.3& 71.2& \textbf{69.7}& 78.2& 66.6& 69.4& \textbf{68.0}\\
      \cline{2-10}

      & ReaLiSe& 78.4&	67.8&	71.5&	{69.6}&	77.7&	66.3&	70.7&	{68.1}\\
      & ReaLiSe-ATLAs& 79.1& 69.1& 75.6& \textbf{72.2}& 78.3& 67.7& 74.0& \textbf{70.7} \\
      \cline{1-10}
      
      \multirow{8}*{SIGHAN 2015} & BERT-Finetune & 83.0&	73.6&	79.5&	{76.4}&	81.8&	71.4&	77.1&	{74.1} \\
      & BERT-ATLAs& 85.5 & 77.8 & 81.4 & \textbf{79.6} & 84.9 & 76.5 & 80.1 & \textbf{77.6} \\
      \cline{2-10}

      & Soft-Masked BERT&83.3&	74.4&	79.9&	{77.0}&	82.1&	72.1&	77.4&	{74.7}\\
      & SMB-ATLAs	& 85.1& 79.0& 80.6& \textbf{79.8}& 84.3& 77.4& 78.9& \textbf{78.1} \\
      \cline{2-10}

      & SpellGCN&84.2&	75.2&	80.2&	77.6&	83.0&	73.0&	77.8&	{75.3}\\
      & SpellGCN-ATLAs& 85.4& 77.4& 82.3& \textbf{79.7}& 84.4& 75.5& 80.2& \textbf{77.8}\\
      \cline{2-10}

      & ReaLiSe&84.7&	77.3&	81.1&	{79.1}&	83.9&	75.8&	79.5&	{77.6}\\
      & ReaLiSe-ATLAs& 86.3& 78.9& 83.5& \textbf{81.1}& 85.7& 77.7& 82.3& \textbf{79.9}\\
      \cline{1-10}
  \end{tabular}
  }
\end{table*}

The accuracy, precision, recall, and F1 score are reported as the evaluation metrics, which are commonly used in the CSC task. All metrics are provided for the detection and correction sub-tasks. All experiments were conducted four times, and the performance of the model using average parameters was reported. The code and the trained model are publicly released at the following address \footnote{\url{https://github.com/1250658183/ATLAs}}.

Table~\ref{table_Results} shows the final performance. In each case, we test the baseline with and without the addition of our ATLAs strategy. In the four baseline models, ATLAs yields consistent performance improvements. Specifically, at the correction level, ATLAs exceeds the sentence level F1 score of the baseline models by 3.1, 2.2, 2.1, 2.6 percent on the SIGHAN 2014, and 3.5, 3.4, 2.5, 2.3 percent on the SIGHAN 2015, which verifies the effectiveness of the ATLAs strategy. It can be observed that ATLAs improves the performance of the vanilla BERT more than the other three baseline models. The reason is that the additional information contained in the layerwise self-attention can help BERT to constrain the similarity between the predicted and the input characters to some extent. The other three baseline models use different modules to model similarity, so there is a partial overlap in the optimization space. Nevertheless, ATLAs is still able to produce stable performance enhancements over the best-performing model ReaLiSe and achieve state-of-the-art results.

We made a more detailed statistical analysis on two baselines. At the sentence level, BERT-Finetune made $585$ adjustments, of which only $418$ adjustments were correct (recorded as $418/585$). Meanwhile, BERT-ATLAs made $424/555$ correct predictions. Similarly, the original ReaLiSe made $433$ correct predictions in $570$ adjustments ($433/570$), while our method made $446/574$ correct predictions. This indicates that ATLAs have the ability to discover more errors and can effectively reduce the overcorrection of the model.

ECOPO~\cite{li2022past} is a related technique that utilizes contrastive probability optimization to adjust the training loss of CSC models. ECOPO combined with ReaLiSe achieves SOTA performance on the SIGHAN datasets by adjusting the robust baseline ReaLiSe. On the SIGHAN 2014 and SIGHAN 2015 datasets, ECOPO combined with ReaLiSe achieves 69.2\% and 78.5\% F1 scores on the correction subtask, respectively. In contrast, our method maintains a performance advantage of 1.5\% and 1.4\% and improves the SOTA performance, further verifying the superiority of our proposed framework.

\subsection{Ablation Study}
We perform ablation experiments on each module to explore the effects of auxiliary task learning based on GMM and layerwise self-attention on the model. The results of ablation experiments are shown in Table \ref{table_Abla}. In group BERT-LA, only the layerwise self-attention strategy is used to fine-tune the BERT-based model. The BERT-ATL group retains only the training strategy based on auxiliary task learning.

\begin{table*}[ht]
  \caption{Results of the ablation experiment (\%). The table shows the results of the experiment on BERT-Finetune performed on SIGHAN 2015.
  }
  \label{table_Abla}
  \centering  
  \setlength{\tabcolsep}{3mm}
  {
  \begin{tabular}{l|c|c|c|c|c|c|c|c}
      \hline
      \makecell[c]{\multirow{2}*{\textbf{Method}}} &\multicolumn{4}{c}{\textbf{Detection-Level}} & \multicolumn{4}{|c}{\textbf{Correction-Level}} \\
      \cline{2-9}  
      & \textbf{Acc.} & \textbf{Prec.} & \textbf{Rec.} & \textbf{F1.} & \textbf{Acc.} & \textbf{Prec.} & \textbf{Rec.} & \textbf{F1.} \\
      \hline
      BERT-Finetune & 83.0&	73.6&	79.5&	{76.4}&	81.8&	71.4&	77.1&	{74.1} \\
      BERT-LA & 84.0& 75.9& 80.4& 78.1& 82.8& 73.6& 78.0& 75.8 \\
      BERT-ATL &84.7&76.7&81.7&79.1&83.2&73.8&78.6&76.1\\
      BERT-ATLAs & 85.5 & 77.8 & 81.4 & {79.6} & 84.9 & 76.5 & 80.1 & {77.6} \\
      \hline
  \end{tabular}
  }
\end{table*}

Layerwise self-attention and GMM-driven auxiliary task provide 1.7 and 2.0 percent improvement on the correction F1 score, respectively. The performance gains from BERT-LA verify that low-level encoding information is beneficial to the CSC task. 
In addition, it can be seen that BERT-ATL can bring about an improvement of 2.7 percent on the detection F1 score. Since the auxiliary task utilizes prior POS information to increase the margin between correct and incorrect character representations during fine-tuning, thus it helps reduce the difficulty of the detection subtask. We show some visualization examples in section \ref{sec_visual}.

\section{Discussion}
\label{sec:discussion}

In this section, we conduct a more in-depth analysis of the effective implementation of each optimization. We explore the results of the experiment on BERT-Finetune performed on SIGHAN 2015.

\subsection{Prior Knowledge Injection Strategy}
To explore the impact of POS information injection strategies on model performance, we compare four different strategies. The results are displayed in Table \ref{table_Abla2}. In Hard-Embedding, the POS feature is concatenated with the BERT hidden state as with DPL-Corr \cite{Xie2019AutomaticCS}. 
Essentially the Hard-Embedding mode directly uses the prior POS as an input feature, DPL-Corr only uses the POS feature for error detection, but we also use it for error correction. In Hard-Joint, $\eta$ is set to 0.5, and $\alpha_i$ is discarded, which has the effect of taking the mean loss of multiple training tasks directly; Full-Annealing uses the loss function proposed in Formula~\ref{equa_Loss:eq2}; Part-Annealing is the combination of the Full-Annealing and Hard-Joint strategies, whereby $\eta$ is calculated as shown in~\eqref{equa_Part_Annealing}.
To further show the effect of error propagation, the column entitled ``C-F1 noise'' displays results when an additional 20\% noise tags are artificially introduced.

\begin{equation}
    \eta_t = \begin{cases}
        \frac{1}{1+e^{\beta(-t+\frac{T}{2})}},\quad t < \frac{T}{2} \\
        0.5, \quad t \geq \frac{T}{2}
    \end{cases}
    \label{equa_Part_Annealing}
\end{equation}

\begin{table}[t]
  \caption{Statistical information of model performance for multiple knowledge injection strategies, where "D-F1/C-F1/C-F1 noise" represents Detection-F1, Correction-F1, Correction-F1 with artificial noise POS labeling, respectively.}
  \label{table_Abla2}
  \centering
\setlength{\tabcolsep}{3mm}
  {
  \begin{tabular}{l|c|c|c}
      \hline
      {\makecell[c]{\textbf{Method}}} &\textbf{D-F1} & \textbf{C-F1} & \textbf{C-F1 noise}  \\
      \hline
      w/o POS         & 78.4 & 75.5 & ---  \\
      Hard-Embedding  & 78.8 & 76.3 & 75.6 \\
      Hard-Joint      & 78.6 & 76.3 & 75.5 \\
      Part-Annealing  & 79.2 & 76.7 & 76.1 \\
      Full-Annealing  & 79.6 & 77.6 & 77.0 \\
      \hline
  \end{tabular}
  }
\end{table}

In columns without artificial noise tags, all the control groups introducing POS information outperformed the base group significantly. This steady improvement confirms that the addition of prior POS tagging is of benefit to the CSC task. 
In addition, annealing strategies consistently yield higher performance than Hard-type strategies because annealing strategies allow the use of GMM and loss annealing to mitigate the impact of noise tags and increase the margin between correct and misspelled character representations, thereby reducing the difficulty of the main CSC task and improving the overall performance.

Moreover, Full-Annealing exhibits better performance than the other three strategies. We believe there are two reasons. First, Full-Annealing does not need to fine-tune the CSC model according to the auxiliary task in the later stage of training. The loss is dominated by the primary correction task, which helps the model converge more smoothly.
Second, the incorrect corpus inevitably introduces noise tags, which would mislead the model using the Hard-type strategy. Full-Annealing utilizes the annealing module weights the auxiliary tags through the probability distribution of auxiliary loss more thoroughly to reduce the impact of noise tags. As shown in the "C-F1 noise" column, Full-Annealing produces a minor performance degradation when knowledge noise is artificially introduced, which also validates this conjecture.

\subsection{Multilayer Representation Strategy} 

To explore the effective utilization strategy of hierarchical knowledge in the CSC model, we carry out the following exploratory experiments. The results are shown in Table \ref{table_Abla3}. In this experiment, in the strategy Mean, the mean of the BERT layer representations is taken; In ResNet, the embedding layer is connected to the top layer via a residual connection; In ResNet5, a uniform selection is made of five BERT internal layers for a residual connection, i.e., layers 3, 6, 9, 12 and the embedding layer; In Last-Query, the query is initiated using the single latent representation of a character in the last hidden layer; Ngram-Query is the strategy defined in \eqref{equa_LA}.

\begin{table}[htbp]
    \caption{Statistical information of model performance for multiple multilayer representation strategies.}
    \label{table_Abla3}
    \centering
    \setlength{\tabcolsep}{3mm}
    {
    \begin{tabular}{l|c|c}
        \hline
        {\makecell[c]{\textbf{Method}}} &\textbf{Detection-F1} & \textbf{Correction-F1} \\
        \hline
        Mean        & 74.9 & 72.0 \\
        ResNet      & 77.6 & 75.8 \\
        ResNet5     & 78.9 & 76.2 \\
        Last-Query  & 79.9 & 77.0 \\
        Ngram-Query   & 79.6 & 77.4 \\
        \hline
    \end{tabular}
    }

\end{table}

As shown in Table \ref{table_Abla3}, the simple Mean does not improve performance relative to baseline but degrades performance, indicating that the naive fusion method is not effective but may introduce a lot of noise to the fused representation. Additionally, the ResNet results in slight performance improvements since the residual connection can help to shrink the margin between the multilayer representation and the target representation for the original correct inputs. When the introduced layers reach ResNet5, the performance improvement tends to be flat, and if all the middle layers are introduced, the model essentially falls back to the Mean policy. Comparing ResNet with layerwise attention, the latter consistently performs better as ResNet is unable to consider the relationship between the different encoder layers, whereas the attention mechanism allows the model to choose information adaptively based on the type of error. 
Ngram-Query yields better performance than the Last-Query strategy. Ngram-Query uses the mixed representation of successive n-gram tokens for query, which can help the model more easily identify whether the input characters are non-word errors, and then help the model allocate attention weight depending on the error type.

\subsection{The Impact of N-gram Size}

To investigate the optimal n-gram size of the layerwise self-attention query, we subsampled a portion of the data and performed comparative experiments. The performance of various n-gram sizes in the error correction subtask is shown in Table \ref{table_ngramsize}. The table findings demonstrate that the shorter unigram and bigram cannot capture the bidirectional contexts of a character, resulting in an inferior performance. The more extended n-gram sizes can cause defocusing in the attention mechanism. Through preliminary experiments, we set the size of the n-gram to 3, which is the smallest size that fuses bidirectional contexts of characters.

\begin{table}[h]
    \caption{Model performance under different n-gram sizes. }
    \label{table_ngramsize}
    \centering
    \setlength{\tabcolsep}{3mm}
    {
    \begin{tabular}{l|cccccc}
        \hline
        \textbf{N-gram} & 1 & 2 & 3 & 4 & 5 & 7 \\
        \hline
        \textbf{Corr. F1} & 75.4 & 75.8 & 76.0 & 75.3 & 74.8 & 74.4 \\
        \hline
    \end{tabular}
    }
\end{table}

\subsection{Comparison of computational Cost}
Quantitative analysis experiments were conducted to assess the impact of ATLAs enhanced framework on model computational complexity, and the results are presented in Table~\ref{table_Abla_add1}. All experiments are conducted on the BERT-Finetune without loss of generality, and analogous conclusions are drawn for other baselines. The model size indicates the local storage size of the model before and after the addition of the ATLAs enhanced framework, the training time indicates the time required to train the model for 1000 rounds with the specified batch size, and the inference time indicates the time required to infer the SIGHAN 2015 test set. The GeForce RTX 3090 graphics card is utilized for all experiments.
The results indicate that the impact of ATLAs on model size is minimal since the EM algorithm does not need to consume local space, and the only source of model size growth is the FC layer used for the auxiliary task. 
In addition, ATLAs increases the training time by 7.7 percent, partly due to the auxiliary task prediction. The more important part comes from the iteration of the EM algorithm at each epoch.  However, it should be noted that the ATLAs framework can maintain the same inference efficacy as the original model since it requires only inference output during the test phase without calculating loss items.

\begin{table}[ht]
    \caption{Quantitative analysis results of model computational complexity}
    \label{table_Abla_add1}
    \centering
    {
    \begin{tabular}{l|c|c}
        \hline
        {\textbf{BERT-Finetune}} &\textbf{w/o ATLAs} & \textbf{w/ ATLAs} \\
        \hline
        Model size         & 392M & 394M \\
        Training time      & 195s & 210s \\
        Inference time     & 3s   & 3s \\
        \hline
    \end{tabular}
    }
\end{table}

\subsection{Visualization Analysis}
\label{sec_visual}
Can the loss annealing strategy effectively inject explicit knowledge?
To get a more intuitive sense of the role of POS information, we performed a t-SNE dimension-reduction analysis\footnote{\url{https://projector.tensorflow.org/}} on the character representations before and after the fine-tuning by auxiliary POS tags. We chose a common character, ``\cjksong{地}''(``\pinyin{de}''), which is a commonly-used auxiliary (-ly) and a noun (ground). By randomly selecting 100 errors and 500 correct instances in the test data, the representations before and after the auxiliary task fine-tuning are embedded as shown in Fig.~\ref{fig_embed_visualization}:

\begin{figure}[ht]
    \centering
    \includegraphics[width=0.38\columnwidth]{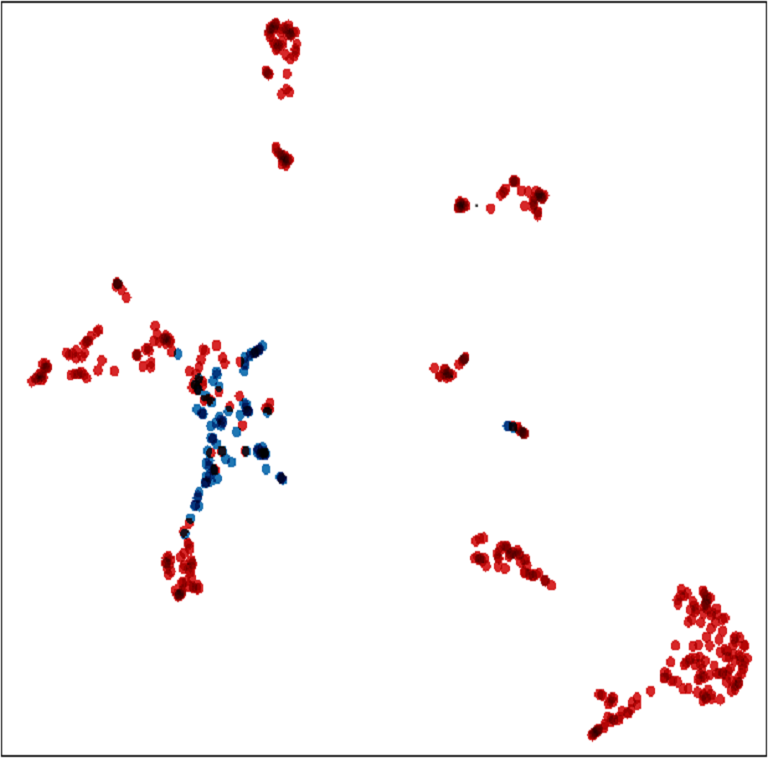} \quad
    \includegraphics[width=0.38\columnwidth]{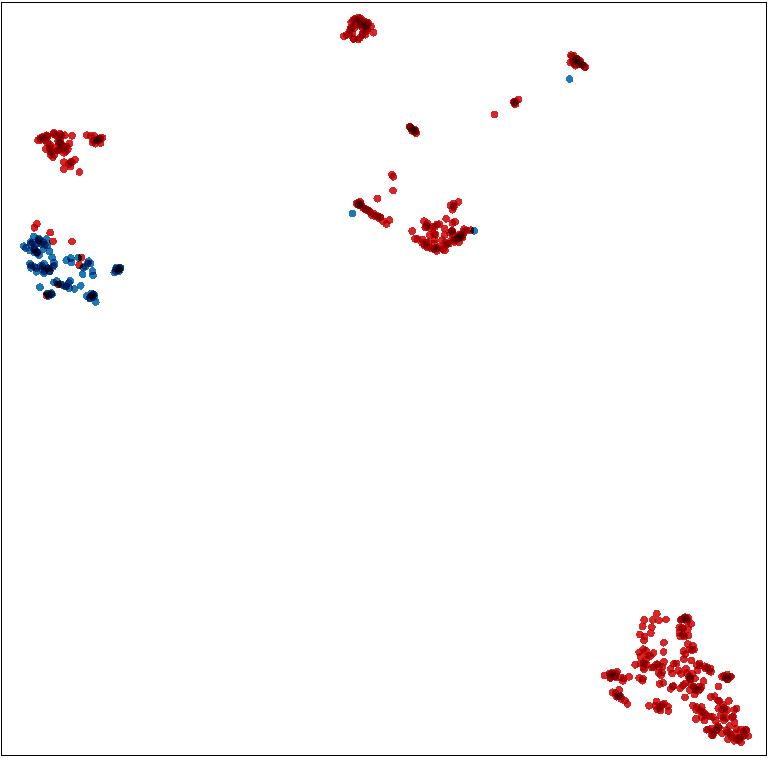}
    \caption{Visualization of character representations via dimension reduction. {\color{red}Red}/{\color{blue}blue} dots correspond to the correct/incorrect characters. The picture on the left displays the representations before information injection, and the picture on the right displays them after injection.}
    \label{fig_embed_visualization}
\end{figure}

The margins between correct and incorrect embeddings are significantly larger after fine-tuning. This shows that POS information can help the model better distinguish between correct and misspelled characters. In addition, after fine-tuning, the degree of aggregation of character embedding is higher because POS information can help the model recognize the usage of different semantics of characters, thus assisting the model in determining whether they are used correctly.

\begin{figure}[t]
  \centering
  \subcaptionbox{Attention map visualization of incorrect character\label{fig_atten_visualization_err}}
  {\includegraphics[width=0.45\columnwidth]{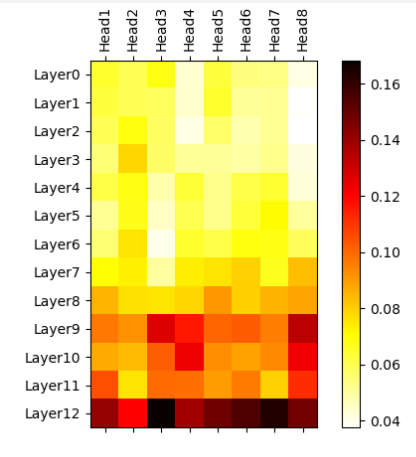}}
  \subcaptionbox{Attention map visualization of correct character\label{fig_atten_visualization_cor}}
  {\includegraphics[width=0.45\columnwidth]{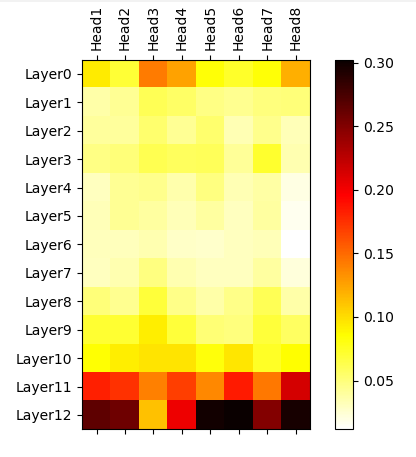}}
  \caption{Visualization of the attention score map of different characters on the intermediary layers.}
  \label{fig_atten_visualization}
\end{figure}

To explore the role of low-level information in correction, we visualized an attention weight map in a set of control instances. Figure \ref{fig_atten_visualization_err} shows the attention map when the input token was modified from ``\cjksong{被坏}'' (be broken) to ``\cjksong{破坏}'' (destroy) in the ``\cjksong{你的工厂把环境被坏}'' (Your factory destroys the environment) example. This instance was not modified in the baseline model. By fusing richer intermediate information, the model can identify and correct this misspelling. Figure \ref{fig_atten_visualization_cor} shows the attention map when the correct ``\cjksong{号}'' (number) is left unchanged in ``\cjksong{我们对号码有兴趣}'' (We are interested in numbers). The baseline model incorrectly changed it to ``\cjksong{数}'' (digital). ATLAs allocates attention to the embedding layer to reduce the distance between the final representation and the original embedding for the original correct inputs. This helps the model predict the original characters, intuitively explaining why the ResNet group can increase model performance.

\subsection{Limitations}

Although the proposed framework obtains superior performance, it has limitations, which are discussed below. First, the existing methods have only been verified on the CSC model, which means our framework currently only serves spelling error correction under equal lengths of input and output. The method's applicability has yet to be explored for Chinese grammatical errors with more complex error types. Second, we conducted a case analysis of the shortcomings of existing methods. We discovered that a considerable number of model errors occurred when multiple or consecutive erroneous characters were in the input sentence. We believe this is because the baseline model predicts the correct characters in a non-autoregressive manner. Therefore, incorporating an optimization strategy that considers context-dependent modeling on top of our method could further improve the error-checking performance of the model.

\section{Conclusion}
\label{sec:conclusion}

In this article, we propose a universal optimization strategy for the BERT-based CSC models based on the fusion of heterogeneous knowledge, including both explicit POS knowledge and implicit hierarchical linguistic information. We propose an annealing loss strategy driven by GMM, which can inject prior information into the BERT representation while reducing the impact of noise tags. We also explore that utilizing intermediary information in the encoder can improve the overall performance of the CSC task, and experimentally verify the effectiveness of the proposed n-gram-based layerwise self-attention mechanism. Experimental results demonstrate that the BERT-based CSC models can be steadily improved after utilizing our proposed ATLAs. We remain to extend the heterogeneous knowledge to the CSC task (e.g., coreference resolution) as our future work.

\section*{Acknowledgements}
This work is supported by the National Natural Science Foundation of China (No. 61936012, 61976114, and 62206126).

\bibliographystyle{IEEEtran}
\bibliography{IEEEabrv,custom}

\end{document}